\documentclass[10pt]{article}
\usepackage[letterpaper, margin=1in]{geometry}
\usepackage{graphicx}
\usepackage{apacite}
\usepackage{natbib}
\usepackage{url}

\usepackage{amsmath}
\usepackage{float}
\usepackage[para]{footmisc}
\usepackage{graphicx}
\usepackage{multirow}
\usepackage{booktabs}
\usepackage{subfig}
\usepackage{enumitem}
\usepackage{wrapfig}
\usepackage[format=plain, labelfont=it, textfont=it, font=small]{caption}

\usepackage[spanish, es-noindentfirst, es-noshorthands, es-tabla]{babel}
\addto{\extrasspanish}{} 
\addto\captionsspanish{\renewcommand{\BOthers}[1]{et al.\hbox{}}}

\usepackage[x11names]{xcolor}
\usepackage[colorlinks, bookmarksopen, bookmarksnumbered, citecolor=Red4, urlcolor=Turquoise4, linkcolor=Green4]{hyperref}

\usepackage{helvet}

\usepackage{sectsty}
\sectionfont{\fontsize{11}{13.2}\selectfont}

\makeatletter
\renewcommand{\maketitle}{\bgroup\setlength{\parindent}{0pt}
    \centering
    \textbf{\@title}
    \begin{flushleft}
        \@author
    \end{flushleft}\egroup
}
\makeatother

\title{\fontsize{18}{21.6}\selectfont \textbf{SISTEMA DE NAVEGACIÓN DE COBERTURA PARA VEHÍCULOS NO HOLONÓMICOS EN AMBIENTES DE EXTERIOR} \par}

\author{
\fontsize{10}{12}\selectfont
\textbf{Michelle Valenzuela$^{1}$, Francisco Leiva$^{1}$, Javier Ruiz-del-Solar$^{1,2}$} \\
$^{1}$\textit{Advanced Mining Technology Center, Universidad de Chile, Chile}\\
$^{2}$\textit{Departamento de Ingeniería Eléctrica, Universidad de Chile, Chile}\\
}

\date{}

\begin{document}

\maketitle
\thispagestyle{empty}

\section*{RESUMEN}
En robótica móvil, la navegación de cobertura hace referencia al desplazamiento deliberado de un robot con el objetivo de cubrir un área o volumen determinado. Ejecutar esta tarea apropiadamente es fundamental para la realización de múltiples actividades, por ejemplo, la limpieza de un recinto por una aspiradora robot. En el campo de la minería, se requiere realizar tareas de navegación de cobertura en distintos procesos unitarios relacionados con el movimiento de material usando maquinaria industrial, como por ejemplo, en tareas de limpieza, en el movimiento de material en botaderos y en la construcción de paredes de tranques de relaves. La automatización de estos procesos es fundamental para mejorar la seguridad asociada a su realización. En este trabajo se presenta un sistema de navegación de cobertura para un robot no holonómico, el cual corresponde a una prueba de concepto para la posible automatización de distintos procesos unitarios que requieren navegación de cobertura, como los mencionados anteriormente. El sistema desarrollado incorpora el cálculo de rutas que permiten a la plataforma móvil cubrir un área determinada, e incorpora maniobras de recuperación en caso de imprevistos frente a la aparición de obstáculos dinámicos o no previamente mapeados en el terreno a cubrir, como otras máquinas o personas que transiten en el sector, pudiendo realizar maniobras de evasión y de recuperación a posteriori, para asegurar la completa cobertura del terreno. El sistema fue puesto a prueba en distintos entornos en simulación y posteriormente probado en ambientes de exterior en el mundo real, obteniéndose resultados cercanos al 90\% de cobertura en la mayoría de los experimentos. Como próxima etapa de desarrollo se realizará el escalamiento de la plataforma utilizada a una máquina o vehículo minero, cuya operación será validada en un entorno real. Algunos de los resultados de las pruebas desarrolladas en el mundo real pueden ser observados en el video disponible en \url{https://youtu.be/gK7_3bK1P5g}.

\section*{INTRODUCCIÓN}

La actividad minera requiere de la realización de múltiples tareas complementarias a la extracción de mineral. Algunas de estas tareas son el manejo de residuos mineros en relaves y botaderos utilizando maquinaria industrial y la mantención de caminos. El manejo de residuos es una tarea crítica que ha adquirido creciente relevancia a medida que aumenta la producción y baja la ley de mineral. Es imperativo realizar una buena gestión de residuos mineros para mantener la seguridad no solo de los trabajadores, sino para asegurar una operación eficiente y sustentable de relaves y botaderos de material. Por otra parte, la mantención de caminos es relevante para mantener altos niveles de producción; los caminos en mal estado derivan en retrasos e incluso posibles accidentes durante el transporte de material, lo cual causa un perjuicio no solo económico, sino también en términos de seguridad e imagen corporativa.

Aunque la gestión de residuos mineros y el mantenimiento de caminos parecen tareas sin nada en común, ambas requieren la operación de maquinaria industrial en un área específica que debe ser cubierta. En la gestión de residuos, sucede en la compactación de relaves y botaderos; mientras que en el mantenimiento de caminos, se da en tareas de nivelación y humectación. En robótica móvil, la tarea de cubrir una cierta área operando una máquina o robot se conoce como \textit{coverage navigation} o navegación de cobertura.

Si bien la navegación de cobertura ha sido ampliamente estudiada, la mayoría de las aplicaciones se enfocan en agricultura y limpieza de interiores, habiendo pocos estudios enfocados en minería. Además, la mayoría de estos sistemas se centran en la planificación y ejecución de rutas, sin considerar mecanismos de recuperación, como la re-planificación para cubrir áreas que quedaron fuera de la cobertura.

Este trabajo presenta un sistema de navegación de cobertura que, además de planificar y ejecutar rutas, incorpora un mecanismo de recuperación para asegurar una mejor cobertura. El sistema también permite que la plataforma móvil evada obstáculos. La implementación presentada corresponde a una prueba de concepto que busca evaluar aspectos clave del problema, identificando componentes críticos, falencias y oportunidades de mejora. 

En una etapa posterior, se espera probar el prototipo en un entorno más cercano a lo que se podría encontrar en un ambiente minero. El desarrollo de un sistema de cobertura que permita la ejecución autónoma de tareas como la construcción de tranques de relaves podría permitir aumentar la seguridad al alejar a las personas de entornos peligrosos y favorecer la diversificación e inclusión laboral al reducir requisitos como licencias especializadas. Además, la integración de este sistema podría incrementar la eficiencia mediante operación continua.
\section*{METODOLOGÍA}

El problema de navegación de cobertura puede considerarse un problema de navegación clásico si las rutas de cobertura son representadas como series de poses consecutivas que una plataforma móvil debe seguir. Dada la importancia del desplazamiento autónomo en aplicaciones robóticas, se han realizado numerosas investigaciones y propuesto estrategias para abordar este problema. El enfoque planteado aplica estos conocimientos al problema de navegación de cobertura de manera sencilla.

El problema de navegación clásico puede resolverse de diversas maneras (e.g.,~\cite{Giralt1990AnIN,Nilsson1969AMA}). Un enfoque común en la literatura es descomponer el problema en subtareas para facilitar su resolución. Una forma de hacerlo es dividirlo en localización, mapeo y planificación global y local. Cada uno de estos componentes aborda una tarea específica, como se detalla a continuación:

\begin{itemize}[noitemsep]
    \item \textbf{Mapeo}: Se refiere a la construcción de una representación del espacio, como mapas de tipo \textit{occupancy grid} o topológicos. Aunque contar con un mapa es fundamental para la navegación, también es posible navegar sin él.
    \item \textbf{Localización}:  Permite a la plataforma móvil determinar su pose en el espacio. La estrategia varía según la disponibilidad de un mapa. Si existe un mapa, la pose se estima comúnmente usando filtros de Kalman (e.g., \cite{yuzhen2016application}) o filtros de partículas (e.g.,~\cite{wolf2005robust}). Sin un mapa previo, el problema se combina con el mapeo, convirtiéndose en un problema de SLAM.
    \item \textbf{Planificación global}: Utilizando una representación del espacio, este componente calcula una ruta hacia un destino, evitando obstáculos en el mapa. Ejemplos de planificadores globales incluyen a A$^{\star}$ \citep{Hart1968AFB} y al algoritmo de Dijkstra \citep{Dijkstra1959ANO}. 
    \item \textbf{Planificación local}: Permite a la plataforma móvil llegar a un destino local a través de información local, evitando obstáculos no identificados en el mapa. Este componente ajusta el plan global según sea necesario y está estrechamente relacionado con las restricciones cinemáticas de la plataforma, que determinan los movimientos que esta puede ejecutar.
\end{itemize}

\begin{figure}[H]
    \centering
    \includegraphics[width=0.9\linewidth]{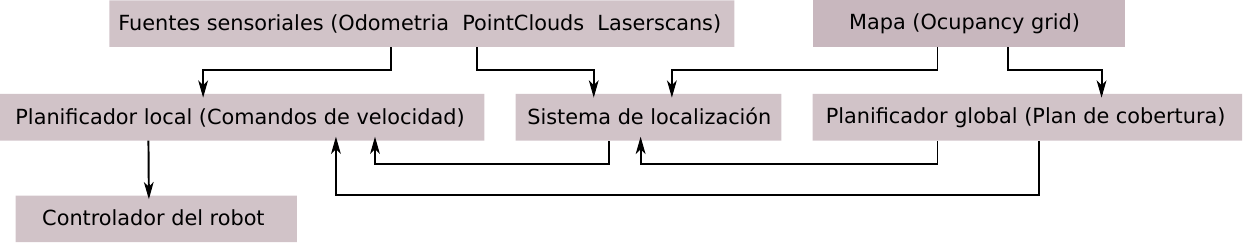}
    \caption{Sistema de navegación de cobertura conceptual. El diagrama ilustra los componentes principales y sus interacciones. Se incluye una componente de fuentes sensoriales, que aunque no forma parte del sistema, es fundamental para la navegación, ya que se necesita información exteroceptiva y propioceptiva para alimentar a otros componentes.}
    \label{fig:nav_stack_min}
\end{figure}

En la navegación de cobertura, el planificador global adquiere un significado diferente; en lugar de calcular una ruta hacia un destino en particular, debe generar una ruta que permita a la base móvil cubrir un área específica del mapa. Estas rutas son obtenidas mediante algoritmos de cálculo de rutas de cobertura. A partir de esto último, y de la descomposición de tareas antes descrita, se propone el sistema de navegación de cobertura conceptual que se ilustra en la Figura~\ref{fig:nav_stack_min}.

La implementación del sistema conceptual propuesto consta de dos partes: por un lado, el desarrollo se enfoca en los algoritmos de cálculo de rutas de cobertura y, por otro, en el sistema de navegación clásico que ejecuta la ruta. Para la componente de cálculo de rutas, se utilizó el trabajo desarrollado por~\cite{bormann2018indoor}, que incluye implementaciones para seis algoritmos de cálculo de rutas de cobertura, los cuales, siguiendo la notación del trabajo original antes señalado, son descritos de manera sucinta a continuación.

\begin{figure}[h]
    \subfloat[\label{fig:dec_exacta}]{\includegraphics[width=0.43\linewidth]{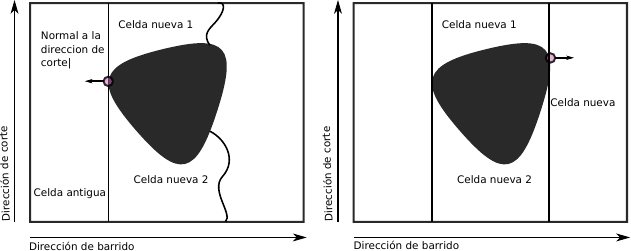}}
    \hfill
    \subfloat[\label{fig:grid_algs}]{\includegraphics[width=0.52\linewidth]{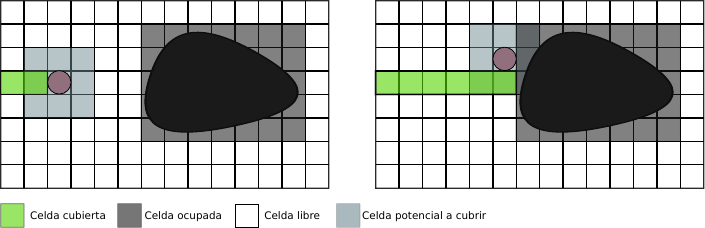}}
    \quad
    \subfloat[\label{fig:voronoi}]{\includegraphics[width=0.28\linewidth]{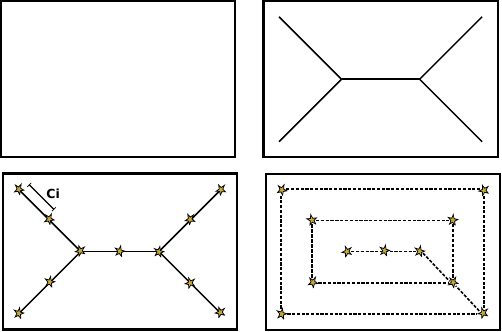}}
    \hfill
    \subfloat[\label{fig:art-gallery}]{\includegraphics[width=0.68\linewidth]{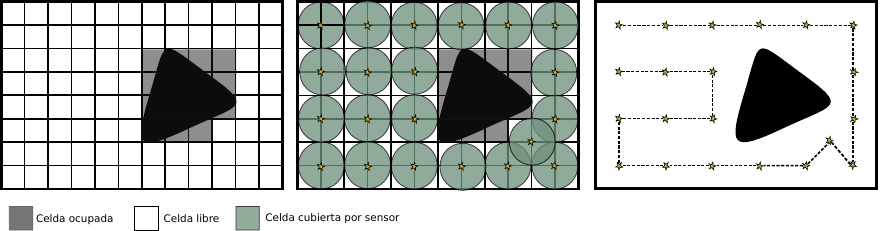}}
    \caption{Algoritmos de planificación de cobertura. (a) Descomposición en celdas de Boustrophedon. (b) Funcionamiento de algoritmos de minimización local de energía y bioinspirados en redes neuronales. (c) Funcionamiento del algoritmo basado en líneas de contorno. (d) Funcionamiento del algoritmo de planificación con sensores convexos.}
    \label{fig:cpp-explain-figs}
    %\vspace{-10pt}
\end{figure}

    \textbf{Descomposición Boustrophedon} \citep{choset1998coverage}: Este algoritmo divide el espacio en celdas que conforman una partición exacta del espacio libre, lo que permite reconstruirlo de manera completa. Para calcular dichas celdas se define una dirección de barrido con la cual se recorre el mapa, identificando puntos críticos que determinan la apertura o cierre de una celda. Estos puntos críticos corresponden a aquellos en que la normal al contorno es perpendicular a una dirección de corte definida, como se ilustra en la Figura~\ref{fig:dec_exacta}. Una vez que el espacio ha sido recorrido y las celdas definidas, se debe establecer el orden de visita de estas. Este problema se aborda formulándolo como una variante del problema del vendedor viajero, en el que cada celda es tratada como un nodo a visitar. Finalmente, la cobertura dentro de cada celda se realiza mediante trayectorias en zigzag de tipo boustrophedon.

    \textbf{Problema del vendedor viajero usando grillas}: Este algoritmo realiza una descomposición aproximada del espacio mediante una cuadrícula uniforme de celdas, como se muestra en la Figura \ref{fig:grid_algs}. Con esta representación, el problema se formula como una variante del problema del vendedor viajero, en el que cada celda corresponde a un nodo a visitar. El tamaño de las celdas está condicionado por las dimensiones del robot, ya que se asume que al ingresar en una celda esta queda completamente cubierta. Finalmente, para calcular la trayectoria entre pares de celdas consecutivas se emplean algoritmos de planificación global, como A$^{\star}$, sobre el mapa original del entorno.

    \begin{wrapfigure}{r}{0.35\textwidth}
    \vspace{-15pt}
    \centering
    \begin{equation}
    \label{eq:redes}
        p_{n} = \max \left\{ x_{j} + c \left(1 - \frac{\Delta \theta_{j}}{\pi}\right) \right\}
    \end{equation}
    \vspace{-20pt}
    \end{wrapfigure}
    
    \textbf{Redes neuronales Bio-inspiradas} \citep{Yang2004ANN}: Este algoritmo particiona el espacio en celdas uniformes mediante una cuadrícula, obteniendo así una descomposición aproximada del área a cubrir. La ruta de cobertura se determina a través de un enfoque bioinspirado en la dinámica de activación neuronal, donde cada celda se modela como una neurona conectada con sus ocho vecinos adyacentes. Es importante notar que este enfoque asume que una vez que el robot entra a la celda esta es cubierta. En cada iteración, para la celda actual se calcula el potencial de cada vecino según~\eqref{eq:redes}, donde $p_{n}$ corresponde a la siguiente celda a visitar, $x_{j}$ representa el vecino $j$-ésimo, $c$ es una constante y $\Delta\theta_{j}$ indica la diferencia angular entre dicho vecino y la posición actual. La celda seleccionada para el siguiente movimiento será aquella con mayor potencial de activación. Para evitar que el robot quede atrapado en zonas estrechas o esquinas, se permite que las celdas sean visitadas más de una vez. El proceso continúa iterativamente hasta que todas las celdas han sido cubiertas.

    \begin{wrapfigure}{r}{0.22\textwidth}
    \vspace{-25pt}
    \centering
    \begin{equation}
    \label{eq:contour}
        c_{i} = r_{c} + 2\cdot i \cdot r_{c}
    \end{equation}
    \vspace{-25pt}
    \end{wrapfigure}
    
    \textbf{Planificación basada en líneas de contorno}: El enfoque de este algoritmo se basa en el uso del esqueleto de Voronoi del espacio a cubrir, sin requerir un particionamiento explícito del área. La estrategia consiste, en primer lugar, en calcular el esqueleto del grafo de Voronoi del entorno. Posteriormente, para cada celda se identifican los puntos críticos, que corresponden a mínimos locales en el esqueleto y suelen aparecer en zonas estrechas o pasajes angostos. Conectando estos puntos con el obstáculo más cercano se obtiene una descomposición similar a la de las celdas de Voronoi. A continuación, en cada celda se computa una transformada de distancia, que asigna a cada punto su distancia al obstáculo o borde más próximo, generando líneas de contorno equidistantes dentro de la celda. Considerando que la cobertura del robot corresponde a $\omega_{c} = 2 \cdot r_{c}$, siendo $r_{c}$ el radio de la circunferencia mínima que inscribe su \textit{footprint} y \eqref{eq:contour}, donde i corresponde al contorno i-esimo, se determinan los niveles de contorno a recorrer. La Figura~\ref{fig:voronoi} ilustra el funcionamiento de este algoritmo.

    \textbf{Planificación mediante colocación de sensores convexos} \citep{Arain2015EfficientMP}: El enfoque tomado por este algoritmo se basa en llevar el problema de cobertura al problema de la galería de arte. Para lograr esto se divide el mapa del área a cubrir en celdas de igual tamaño y se define el área de cobertura del robot como el FoV del sensor. Se asume que si una celda es visible por este FoV es cubierta por el robot. Resolver el problema de esta manera permite encontrar el número mínimo de poses a las que la plataforma móvil debe viajar para cubrir el área de interés. Para determinar el mejor orden de visita de estas poses se debe resolver el problema del vendedor viajero, donde cada pose es un pueblo. Lo anterior es resumido en el caso de ejemplo que se presenta en la Fig.~\ref{fig:art-gallery}. 

    \begin{wrapfigure}{r}{0.43\textwidth}
    \vspace{-20pt}
    \centering
    \begin{align}
        E(p, n) &= d_{t}(p, n) + d_{r}(p, n) + N(n) \label{eq:cost} \\
        d_{t}(p, n) &= \frac{\sqrt{(p_{x} - n_{x})^2 + (p_{y} - n_{y})^2 }}{l} \label{eq:d-xy}\\
        d_{r} &=  \frac{| p_{\theta} -n_{\theta} |}{\pi/2}  \label{eq:d-theta}\\
        N(n) &= 4 -\sum_{k\in Nb_{8}(n)} \frac{|k \cap L|}{2} \label{eq:atract}
    \end{align}
    \vspace{-10pt}
    \end{wrapfigure}
    
    \textbf{Minimización local de energía basada en grilla} \citep{Bormann2015NewBS}: El enfoque de este algoritmo se basa en la división uniforme del espacio a cubrir, determinando el orden de visita de las celdas mediante la minimización de una función de energía. Al igual que en otros métodos que emplean una discretización en grillas, se asume que una celda queda cubierta una vez que el robot ha ingresado en ella. El orden de visita se establece mediante la función de costo definida en~\eqref{eq:cost}. En cada iteración, la siguiente celda $n$ se selecciona entre los ocho vecinos de la celda actual $p$. La función de energía \eqref{eq:cost} incorpora en el costo la distancia traslacional (Ecuación~\eqref{eq:d-xy}), la distancia rotacional (Ecuación~\eqref{eq:d-theta}) y un término atractivo $N(n)$ (Ecuación~\eqref{eq:atract}), que incentiva al robot a preferir aquellas celdas cercanas a las ya cubiertas. En esta formulación, $L$ representa el conjunto de todas las celdas visitadas y $Nb_{8}(n)$ el conjunto de los ocho vecinos de la celda $n$. 

\begin{figure}[h]
    \centering
    \subfloat[\label{fig:stack-mb}]{\includegraphics[width=1\linewidth]{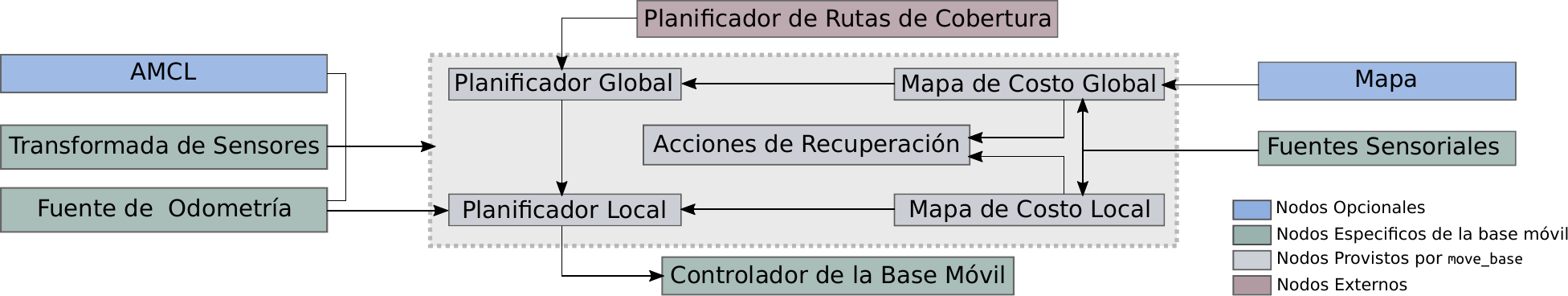}}
    \hfill
    \subfloat[\label{fig:stack-rl}]{\includegraphics[width=1\linewidth]{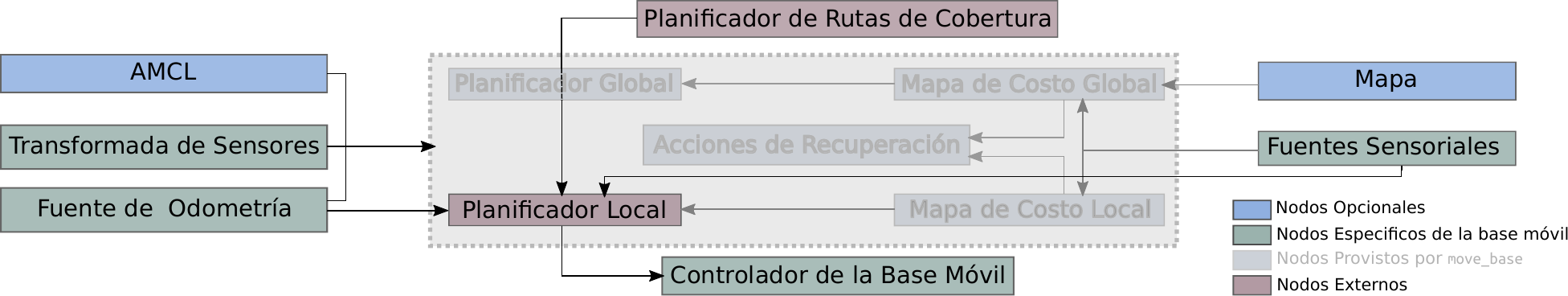}}
    \caption{Sistemas implementados. (a) Usando \texttt{move\_base}. (b) Usando planificador local basado en RL.}
    \label{fig:implementacxiones-stacks}
\end{figure}

Para el sistema de navegación clásico se implementaron dos versiones. La primera utiliza \texttt{move\_base}\footnote{\url{http://wiki.ros.org/move_base}} de ROS, requiriendo ajustes en los parámetros de sus componentes para integrarse a la plataforma móvil. La segunda utiliza el planificador local propuesto en~\cite{leiva2024combining}, el cual se entrena usando aprendizaje reforzado (RL). La Figura \ref{fig:implementacxiones-stacks} ilustra los componentes específicos de cada sistema y sus interacciones. Como se observa, el segundo sistema es más simple, ya que requiere menos componentes y no necesita un planificador global intermedio (distinto a los de planificación de cobertura, por ejemplo,~A$^{\star}$).

\begin{figure}[h]
    \centering
    \includegraphics[width=1\linewidth]{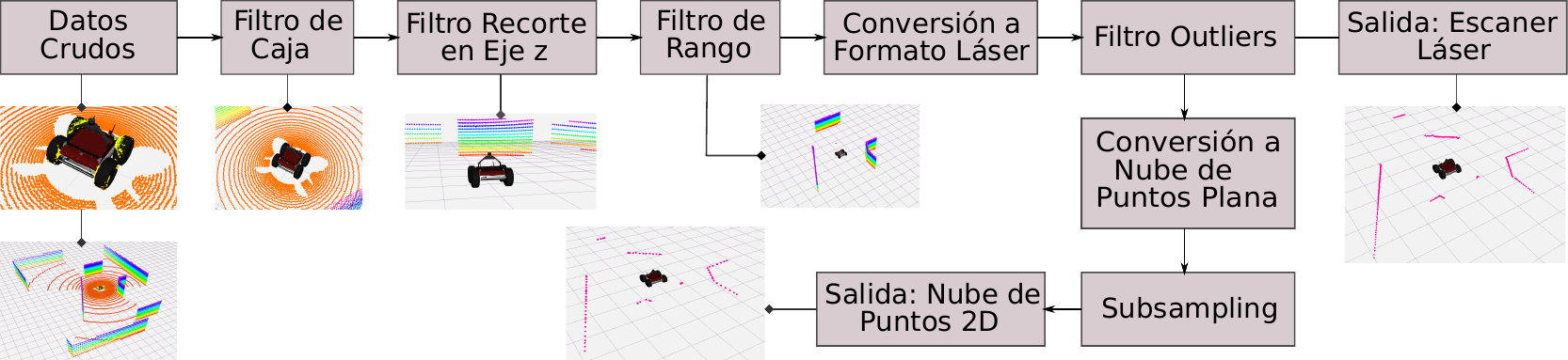}
    \caption{Pipeline de aplicación de filtros y transformaciones del nodo de procesamiento de fuentes sensoriales. El dato de entrada (dato crudo) corresponde a una nube de puntos tridimensional.}
    \label{fig:ouster-filters}
    %\vspace{-10pt}
\end{figure}

El sistema de localización utilizado en ambos sistemas es AMCL \citep{dellaert1999monte}, usando la implementación de ROS. Para el mapeo, se utilizó \texttt{gmapping} \citep{Grisetti2007ImprovedTF}, también usando la implementación de ROS. En cuanto a las fuentes sensoriales, se desarrolló un nodo externo que procesa datos de un sensor con formato de nube de puntos tridimensional, filtrando información irrelevante para la navegación, como obstáculos del suelo, del cuerpo de la plataforma y puntos espurios (\textit{outliers}). Este nodo genera dos formatos de salida: uno en formato escáner láser, que alimenta todos los componentes del sistema, excepto el planificador local basado en aprendizaje reforzado, que requiere nubes de puntos planas (2D). El pipeline de filtros y transformaciones de este nodo se muestra en la Figura \ref{fig:ouster-filters}, donde se incluyen imágenes de ejemplo por etapa para ilustrar la función de cada filtro. Adicionalmente, se implementó un \textit{raycasting}~2D que permite delimitar de manera virtual los límites del área a cubrir, generando que estos límites virtuales produzcan mediciones de entrada a los planificadores locales, como si fuesen obstáculos. Esto permite que el robot no salga del área mientras maniobra o evade obstáculos, ya que esto podría ser un comportamiento poco deseable, por ejemplo, por motivos de seguridad.

Para el sistema de navegación basado en \texttt{move\_base}, el planificador global utilizado fue NavFn\footnote{\url{http://wiki.ros.org/navfn}} y el planificador local fue DWA~\citep{Fox1997TheDW}. Para los mapas de costo se utilizó una capa de inflación\footnote{\url{http://wiki.ros.org/costmap_2d}} y se ajustaron los parámetros que provee para adecuarlos a la base móvil usada. Para las acciones de recuperación, se utilizaron las que \texttt{move\_base} provee por defecto. En el caso del sistema de navegación basado en el planificador desarrollado en base a aprendizaje reforzado, estas componentes no son requeridas, a excepción del planificador local propiamente tal (el controlador), cuya implementación es la especificada en~\cite{leiva2024combining}.

Para unificar los planificadores de rutas de cobertura y el sistema de navegación, se requiere una estructura que coordine las tareas de planificación y ejecución secuencial de poses. Además, es necesario incorporar y coordinar un mecanismo de recuperación para realizar re-planificaciones y cubrir áreas no alcanzadas durante la ejecución, lo que puede suceder por obstáculos no identificados en el mapa (e.g., personas u otras máquinas).

\begin{figure}[h]
    \centering
    \includegraphics[width=0.95\linewidth]{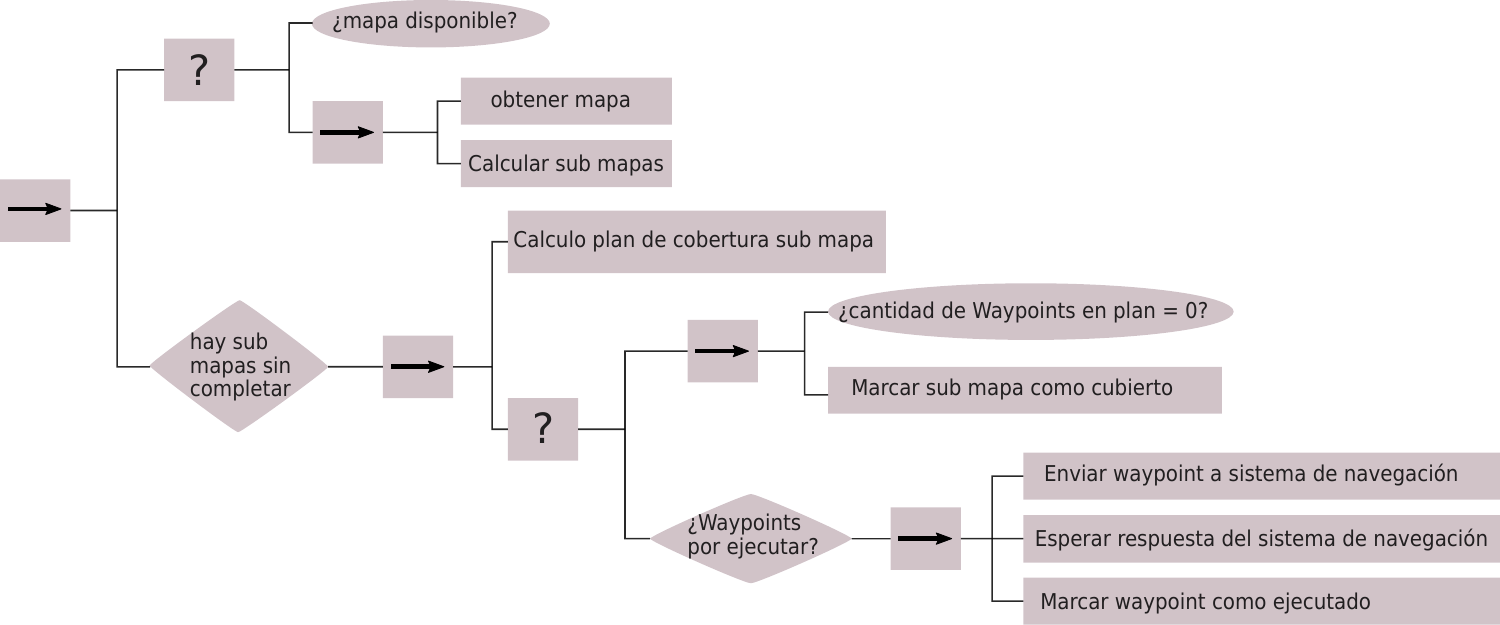}
    \caption{Árbol de comportamiento del sistema de cobertura. El orden de lectura es izquierda-derecha y arriba-abajo.}
    \label{fig:bt-logic}
    \vspace{-10pt}
\end{figure}

Con esto en mente, se diseña un árbol de comportamiento \citep{Colledanchise2017BehaviorTI}, cuya lógica se ilustra en la Figura \ref{fig:bt-logic}. Este árbol divide el área a cubrir en sub-áreas más pequeñas, realizando chequeos después de ejecutar el plan de cobertura para cada una. Se asume que si el planificador no entrega un nuevo plan, el área ya está cubierta, permitiendo avanzar a la siguiente. Así, es posible cubrir zonas que fueron omitidas debido a obstáculos que interrumpieron el plan de cobertura. Adicionalmente, el realizar estas sub divisiones permite volver a cubrir estas zonas más rápido que si esto se realizara al final de la ejecución y también podría facilitar la posibilidad de tener múltiples robots ejecutando la tarea, al poder asignar áreas de trabajo.

\section*{RESULTADOS}

La evaluación del sistema fue dividida en dos aristas principales. La primera busca evaluar y comparar los algoritmos de planificación de rutas de cobertura. La segunda busca evaluar el sistema de cobertura, realizando una comparación entre los sistemas de navegación implementados, haciendo diferenciación respecto a los planificadores locales utilizados, i.e., DWA y basado en aprendizaje reforzado (RL), que son los que ejecutan la ruta calculada y por ende su desempeño tiene un impacto importante en el resultado.

\subsection*{\textit{Benchmark} de algoritmos de cálculo de rutas de cobertura}

El cálculo de rutas de cobertura es fundamental para el sistema, ya que lo diferencia de un sistema de navegación clásico. Por ello, se evaluó el comportamiento de los algoritmos generadores de rutas utilizando varias de las métricas propuestas en~\cite{bormann2018indoor}, y que son descritas a continuación.

    \begin{wrapfigure}{r}{0.2\textwidth}
    \vspace{-20pt}
    \centering
    \begin{equation}
         T_{\text{c}} = T_{f} - T_{0}
        \label{eq:t-calculo}
    \end{equation}
    \vspace{-25pt}
    \end{wrapfigure}
    
    \textbf{Tiempo de cálculo} [$T_{c}$]:  Se define como el tiempo que requiere el algoritmo para calcular el plan de cobertura. Su cómputo se realiza restando el tiempo en el cual el algoritmo termina de realizar el cálculo ($T_{f}$) y el tiempo en el cual comienza ($T_{0}$), de acuerdo a~\eqref{eq:t-calculo}. Este parámetro es importante, pues mientras menor sea el tiempo que se requiere para calcular el plan, más rápido es posible hacer re-planificaciones.

    \begin{wrapfigure}{r}{0.22\textwidth}
    \vspace{-20pt}
    \centering
    \begin{equation}
    \label{eq:coverage-predicho}
        C_{\text{\%}} = \frac{P_{\text{final}}}{P_{\text{inicial}}} \cdot 100
    \end{equation}
    \vspace{-20pt}
    \end{wrapfigure}
    
     \textbf{Cobertura \textit{a priori}} [$C_{\%}$]:  Corresponde al porcentaje del área  de interés que se logra cubrir con el plan entregado, asumiendo que el robot cubre cada pose de este.  El cómputo de esta métrica se realiza calculando la razón porcentual entre los píxeles que no pudieron ser cubiertos por el \textit{footprint} del robot por alguna pose del plan de cobertura, ($P_\text{final}$), y los píxeles del mapa sin cubrir inicialmente, ($P_\text{inicial}$), como se indica en la Ecuación~\eqref{eq:coverage-predicho}. Se dice que es la cobertura \textit{a priori} ya que el robot podría no cubrir todas las poses durante ejecución del plan por distintos motivos, como la presencia de obstáculos dinámicos, o que la pose esté en un lugar no alcanzable por el robot. Esta métrica permite conocer la eficacia de los algoritmos pre-ejecución.

    \begin{wrapfigure}{r}{0.28\textwidth}
    \vspace{-20pt}
    \centering
    \begin{equation}
    \label{eq:largo-cpp}
        P_{\text{len}} = \sum d_{A^{\star}}(W_{i+1}, W_{i})
    \end{equation}
    \vspace{-20pt}
    \end{wrapfigure}

    \textbf{Largo del plan} [$P_\text{len}$]: Es la distancia total que el robot debe viajar, como mínimo, para poder alcanzar todas las poses del plan de cobertura. El cómputo de esta métrica se realiza según~\eqref{eq:largo-cpp}, donde para realizar el cálculo de la mínima distancia entre poses consecutivas, $W_{i}$ y $W_{i+1}$, se utilizó el algoritmo A$^\star$. Esta métrica permite saber qué tan ``óptimo'' es un plan, en el sentido de que a un mismo porcentaje de cobertura, el plan más corto representa una mejor alternativa, ya que el robot debe recorrer menos distancia y por ende demora menos tiempo en alcanzar un mismo nivel de cobertura que un plan más largo.    

    \begin{wrapfigure}{r}{0.25\textwidth}
    \vspace{-20pt}
    \centering
    \begin{equation}
    \label{eq:giros}
        \theta_{\text{T}} = \sum |\theta_{i}| - |\theta_{i+1}|
    \end{equation}
    \vspace{-20pt}
    \end{wrapfigure}

    \textbf{Numero de vueltas} [$\theta_{T}$]:  Esta métrica indica cuántos cambios angulares la máquina deberá efectuar para poder ejecutar el plan de cobertura. Su cómputo se realiza de acuerdo a la Ecuación~\eqref{eq:giros}. Para dos poses consecutivas, $W_{i}$ y $W_{i+1}$, se calcula la diferencia entre sus ángulos, $\theta_{i}$ y $\theta_{i+1}$, y esto se realiza iterativamente sobre todas las poses del plan. Conocer la cantidad de giros que el robot debe realizar puede ser importante por distintos motivos, por ejemplo, en el caso de plataformas de tipo \textit{skid-steered}, los movimientos de giro producen derrape, lo cual puede afectar negativamente a la localización del robot, y en consecuencia puede mermar la calidad de la cobertura. Otro motivo por el cual se podría preferir tener una menor cantidad de movimientos de giro es por la limitada maniobrabilidad que ciertas máquinas pueden tener.

    \begin{wrapfigure}{r}{0.32\textwidth}
    \vspace{-10pt}
    \centering
    \begin{equation}
    \label{eq:densidad-wp}
        \lambda_{\text{path}} = \frac{\sum d(W_{i+1}, W_{i})}{N_{\text{waypoints}}}\frac{1}{2R_{\text{c}}}
    \end{equation}
    \vspace{-20pt}
    \end{wrapfigure}

    \textbf{Separación de poses} [$\lambda_\text{path}$]: Esta métrica indica que tan separados se encuentran, en promedio, poses consecutivas, en relación al tamaño del robot. La importancia de este parámetro radica en que, durante pruebas experimentales, tanto en simulación como en el mundo real, se pueden observar distintos comportamientos según la densidad de poses de un plan. Esta métrica se calcula según la Ecuación~\eqref{eq:densidad-wp}; se computa la distancia promedio entre poses consecutivas, $W_{i}$ y $W_{i+1}$, y se normaliza por el diámetro de la circunferencia mínima ($2R_{c}$) en la cual el robot puede ser inscrito.

\begin{figure}[h]
    \centering
    \includegraphics[width=0.99\linewidth]{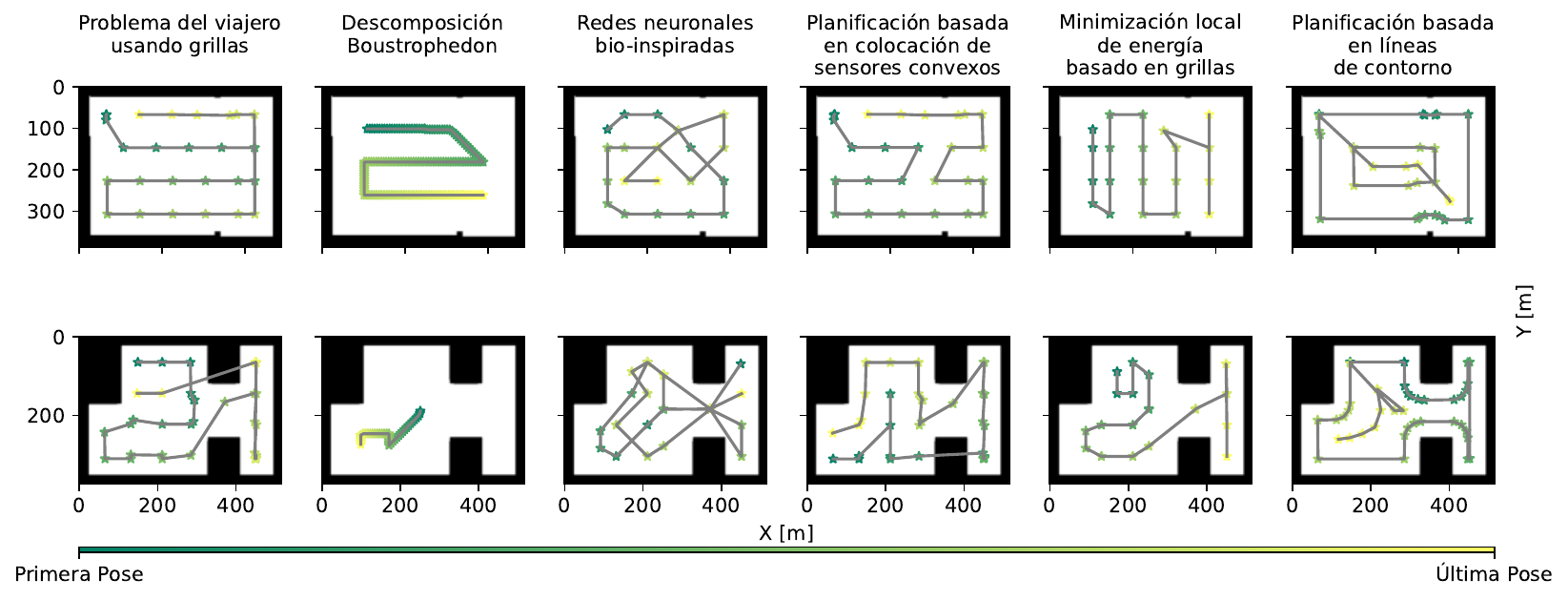}
    \caption{Resultados cualitativos del \textit{benchmark} de algoritmos de planificación de cobertura.}
    \label{fig:benchmark-preview}
    \vspace{-5pt}
\end{figure}

Para la evaluación, se utilizó un subconjunto de mapas del conjunto de datos HouseExpo~\citep{Li2019HouseExpoAL}, abarcando aproximadamente 1000 mapas con tamaños entre 500$\times$500 y 700$\times$700 píxeles. Para cada mapa, se calculó un plan de cobertura por algoritmo y se evaluaron las métricas previamente descritas. Los resultados se resumen en la Tabla \ref{tab:benchmark}, mientras que la Figura \ref{fig:benchmark-preview} muestra resultados cualitativos que evidencian las diferencias en los planes generados por los diferentes algoritmos.

\begin{table}[h]
\caption{Resumen de los resultados obtenidos para el \textit{benchmark} de algoritmos de planificación de cobertura.}
\label{tab:benchmark}
\resizebox{\columnwidth}{!}{
\begin{tabular}{lccccc}
\toprule
\textbf{Algoritmo}&  \textbf{$C_{\%}$ [\%]} & $T$\textbf{$_{c}$ [seg]} & $P_{\text{len}}$ \textbf{[m]} & \textbf{$\theta_{T}$ [rad]} & $\lambda_{\text{path}}$\\ \midrule
Problema del viajero usando grillas & 98.93 $\pm$ 1.29 & 3.47 $\pm$ 3.82 & 56.05 $\pm$ 35.47 & 68.01 $\pm$ 31.01 & 1.30 $\pm$ 0.10 \\ %\midrule
Descomposición Boustrophedon & 86.56 $\pm$ 5.57 & 1.34 $\pm$ 0.73 & 54.25 $\pm$ 35.24 & 29.89 $\pm$ 12.21 & 0.19 $\pm$ 0.02 \\ %\midrule
Redes neuronales bio-inspiradas & 96.04 $\pm$ 2.65 & 0.03 $\pm$ 0.03 & 205.12 $\pm$ 174.65 & 57.94 $\pm$ 52.51 & 1.46 $\pm$ 0.09  \\ %\midrule
P. basada en colocación de sensores convexos & 98.93 $\pm$ 0.01 & 3.37 $\pm$ 3.75 & 61.80 $\pm$ 33.82 & 67.39 $\pm$ 30.82 & 1.30 $\pm$ 0.07 \\ %\midrule
Minimización local de energía basada en grilla & 96.27 $\pm$ 2.54 & 0.008 $\pm$ 0.004 &  75.88 $\pm$ 49.12  & 38.51 $\pm$ 14.69 & 1.45 $\pm$ 0.10 \\ %\midrule
Planificación basada en lineas de contorno & $\,\,$ 62.72 $\pm$ 11.16 & 0.06 $\pm$ 0.02 & 69.35 $\pm$ 36.55 & 52.01 $\pm$ 19.12 & 2.22 $\pm$ 0.85 \\
\bottomrule
\end{tabular}}
\end{table}

De los resultados se observa que la mayoría de los algoritmos lograron un buen porcentaje de cobertura promedio. Sin embargo, los algoritmos de descomposición Boustrophedon y el basado en líneas de contorno obtuvieron resultados considerablemente más bajos, siendo este último el de peor desempeño. Esta baja cobertura se debe a que las poses del plan están demasiado separadas, como también lo refleja la métrica $\lambda_\text{path}$. No obstante, si el robot siguiera la línea punteada de flujo mostrada en las imágenes, se esperaría lograr una cobertura mucho mayor. Cabe destacar que esto depende netamente del planificador local, que es el que determina como moverse de una pose a otra y por tanto la métrica de cobertura, en este caso, solo considera las poses entregadas.

En el caso del algoritmo Boustrophedon, se observa que los planes calculados son más cortos que los de otros algoritmos, siendo insuficientes para cubrir el área completa. Un comportamiento interesante de estos planes es su simplicidad, consistiendo principalmente en líneas rectas a lo largo del lado más largo del mapa, lo que reduce la cantidad de giros que el robot debe realizar, lo que se refleja en la métrica de giro $\theta_{T}$. Además, este algoritmo presenta una densidad significativamente mayor en comparación con los demás. Por ello, se evaluará cómo estas características del plan afectan su ejecución.

Para el resto de los algoritmos, se observa una cobertura promedio superior al 95\%. La principal diferencia entre ellos radica en el tiempo de cálculo promedio para obtener el plan, con una diferencia de orden de magnitud entre los algoritmos basados en el problema del viajero y planificación con sensores convexos, frente a los basados en líneas de contorno, minimización de energía y redes neuronales bio-inspiradas.

En cuanto a la forma del plan (largo y rotaciones), los generados por el algoritmo de redes neuronales bio-inspiradas son notablemente más largos que el resto, los cuales presentan valores similares entre sí. En cuanto al número de giros, el algoritmo de minimización local de energía muestra el menor valor, lo que sugiere que sus planes son más simples, con menos cambios de dirección y una menor cantidad de movimientos en zigzag. Respecto a la densidad de las poses, se observan valores similares entre los algoritmos (salvo Boustrophedon y líneas de contorno), con una separación promedio de aproximadamente 1.38 veces el tamaño del robot.

\subsection*{Evaluación del sistema de cobertura}

\begin{wrapfigure}{r}{0.45\linewidth}
\vspace{-20pt}
\begin{center}
    \includegraphics[width=0.9\linewidth]{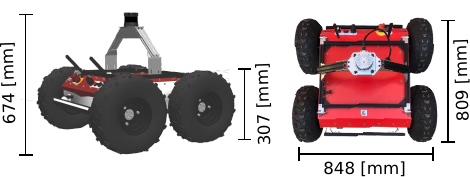}
\end{center}
\vspace{-10pt}
\caption{Plataforma de despliegue Panther con sensor Ouster OS0 montado. La configuración utilizada en simulación y mundo real son equivalentes.}
\vspace{-10pt}
\label{fig:panther_setup}
\end{wrapfigure}

Para evaluar el sistema de cobertura desarrollado, este se desplegó en la plataforma móvil Panther de Husarion. A esta plataforma se le integró un LiDAR Ouster OS0, el cual cuenta con un amplio campo de visión, con 360° de FoV en horizontal, 90° en horizontal y 35 metros de rango. La plataforma, el sensor y su montaje pueden apreciarse en la Figura \ref{fig:panther_setup}.

\begin{figure}[h]
    \centering
    \subfloat[\label{fig:map_simple_1}]{
    \includegraphics[height=0.12\linewidth, trim={1cm 0cm 3cm 0cm},clip]{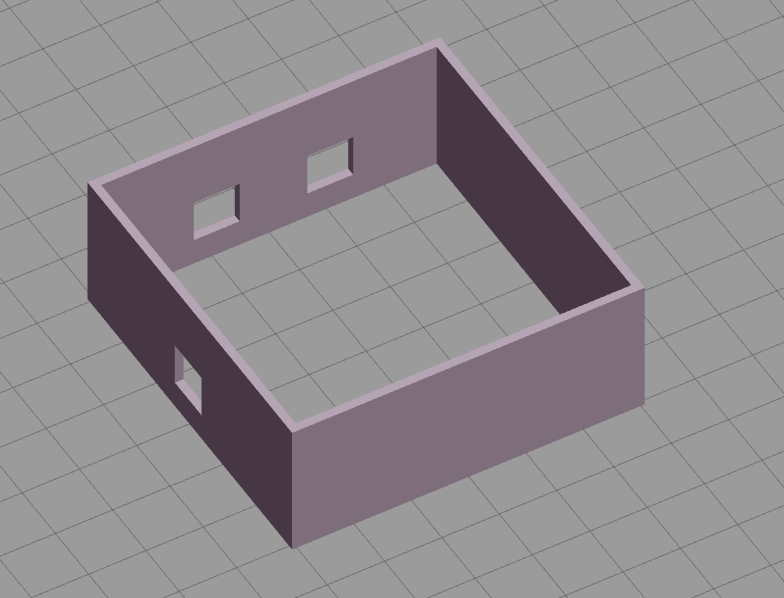}}
    \subfloat[\label{fig:map_simple_2}]{
    \includegraphics[height=0.12\linewidth]{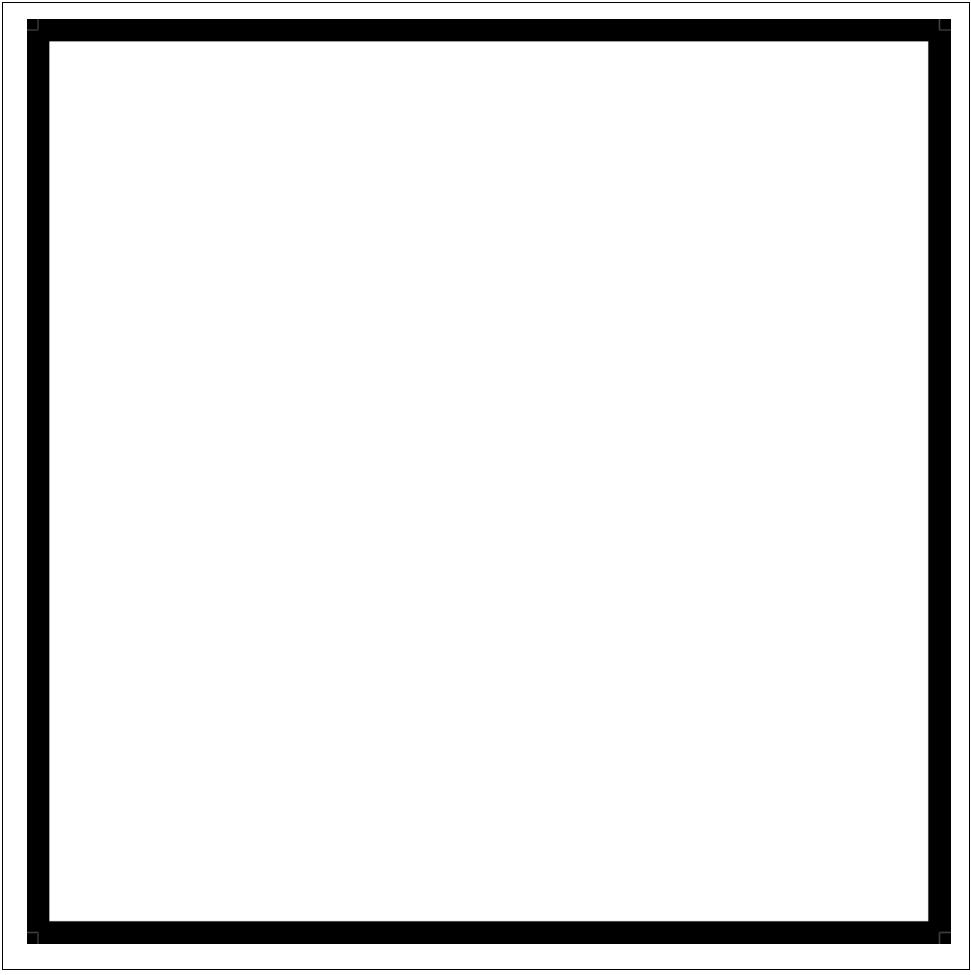}}
    \subfloat[\label{fig:map_obst_1}]{
    \includegraphics[height=0.12\linewidth]{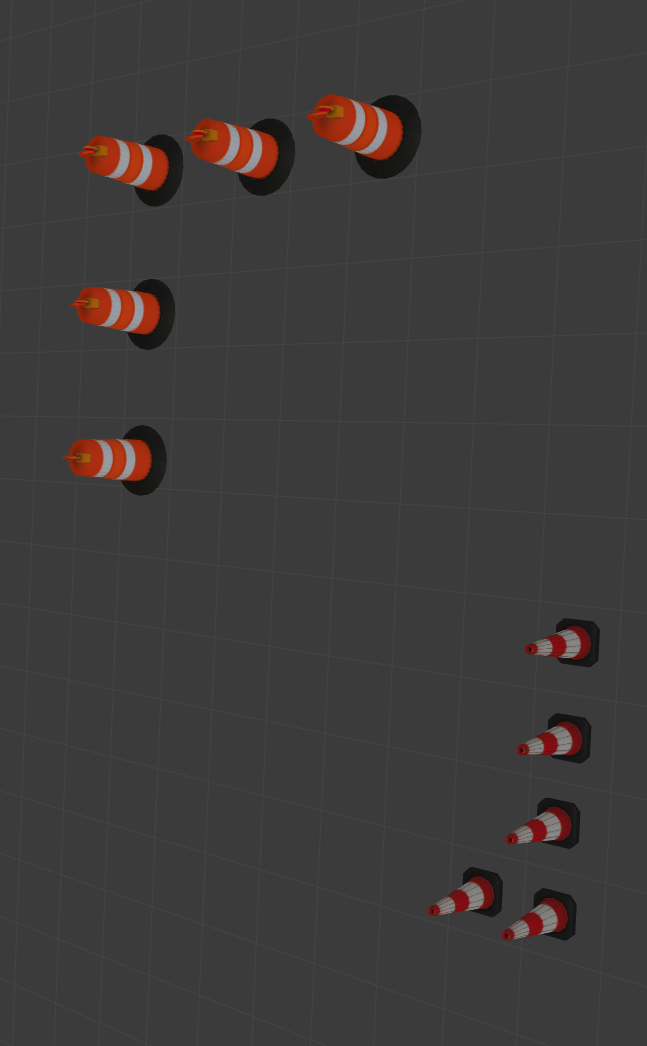}}
    \subfloat[\label{fig:map_obst_2}]{
    \includegraphics[height=0.12\linewidth, trim={5cm 5cm 8cm 7cm},clip]{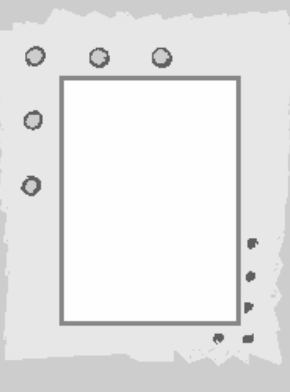}}
    \subfloat[\label{fig:map_real}]{
    \includegraphics[height=0.12\linewidth]{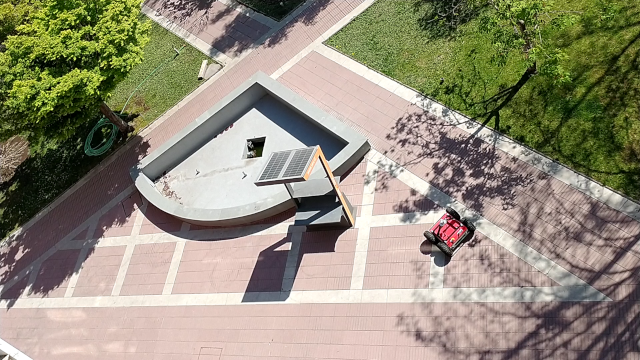}}
    \subfloat[\label{fig:slice_pileta}]{
    \includegraphics[height=0.12\linewidth, trim={9cm 10cm 8cm 9cm},clip]{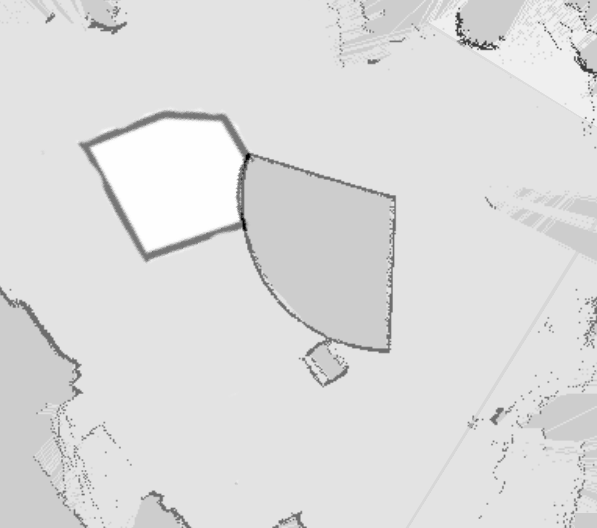}}
    \subfloat[\label{fig:map_pileta}]{
    \includegraphics[height=0.12\linewidth, trim={9cm 10cm 10cm 10cm},clip]{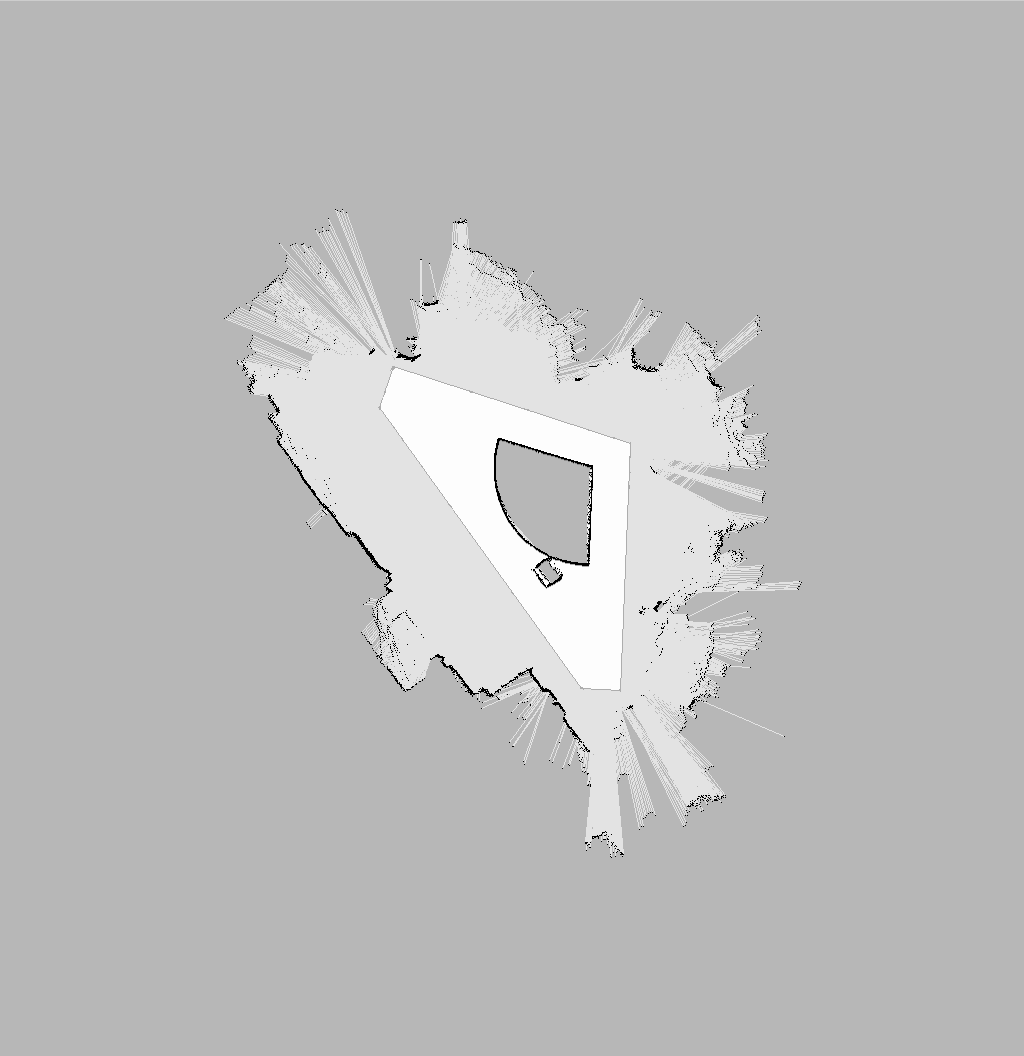}}
    \caption{Escenarios de prueba. (a) Escenario simple en simulación. (b) Mapa del escenario simple, con sección a cubrir. (c) Escenario de prueba en simulación con obstáculo no mapeado. (d) Mapa escenario del simulación con obstáculo con área a cubrir. (e) Escenario de pruebas del mundo real (pileta AMTC).  (f) Sección a cubrir experimento simple en mundo real. (g) Mapa y sección a cubrir en experimento de circuito en torno a pileta.}
    \label{fig:stage-pruebas}

\end{figure}

Los ambientes de prueba utilizados se muestran en la Figura \ref{fig:stage-pruebas}. Es importante notar que se trabaja una sección especifica del área, la cual se encuentra delimitada mediante software (\textit{raycasting}), es decir, no hay barreras físicas reales como vallas o paredes.

Las pruebas en simulación incluyeron una prueba simple inicial y otra con un obstáculo persistente. En las pruebas del mundo real se realizó una prueba simple similar a la de simulación, además de una prueba en un entorno más complejo, sin utilizar obstáculos no identificados. En todos los casos se mide el tiempo de ejecución de la tarea en minutos, $T_\text{exec}$, y el porcentaje real de cobertura alcanzado, $C_{R\%}$, medido de forma análoga a $C_{\text{\%}}$.
\vspace{-10pt}

\subsubsection*{Pruebas en simulación}
\vspace{-5pt}
\begin{wraptable}{r}{0.7\linewidth}
\vspace{-15pt}
\caption{Resultados de prueba en simulación en escenario simple en simulación.}
\label{tab:cuadrado-simu}
\centering
\small
\resizebox{0.7\columnwidth}{!}{
\begin{tabular}{lcccc}
\toprule
\multirow{2}{*}{\textbf{Algoritmo}}&  \multicolumn{2}{c}{\textbf{RL}} & \multicolumn{2}{c}{\textbf{DWA}}  \\ \cmidrule(r){2-5}
 & \textbf{$C_{R\%}$ [\%]} & $T_{\text{exec}}$ \textbf{[min]} & \textbf{$C_{R\%}$ [\%]} & $T_{\text{exec}}$ \textbf{[min]} \\ \midrule
Problema del viajero usando grillas & 95.90 & 5.24  & 95.07 & 24.58\\
Descomposición Boustrophedon & 90.22 & 4.18 & 84.86 & 10.15\\
Redes neuronales bio-inspiradas & 95.73 & 7.01 & 94.59 & 17.43\\
P. basada en colocación de sensores convexos & 95.07 & 5.21 & 94.90 & 24.10 \\
Minimización local de energía basada en grilla & 94.53 & 5.47 & 93.73 & 21.04 \\
Planificación basada en lineas de contorno & 89.18 & 7.58 & 92.42 & 18.50 \\
\bottomrule
\end{tabular} }
\end{wraptable}

La primera prueba se realizó en el entorno simple mostrado en las Figuras~\ref{fig:map_simple_1} y~\ref{fig:map_simple_2}, para evaluar el desempeño del sistema en un caso básico. Los resultados de la Tabla~\ref{tab:cuadrado-simu} muestran que el sistema cumple exitosamente su tarea, alcanzando porcentajes de cobertura superiores al 90\% en la mayoría de los casos. En cuanto al tiempo necesario para completar la tarea, se observa que la ejecución con el sistema de navegación \texttt{move\_base} (DWA) es significativamente más lenta que con el sistema que usa el planificador local entrenado con aprendizaje reforzado, tardando entre 2 y 3 veces más.

\begin{figure}[H]
\vspace{-5pt}
    \centering
    \includegraphics[width=0.99\linewidth]{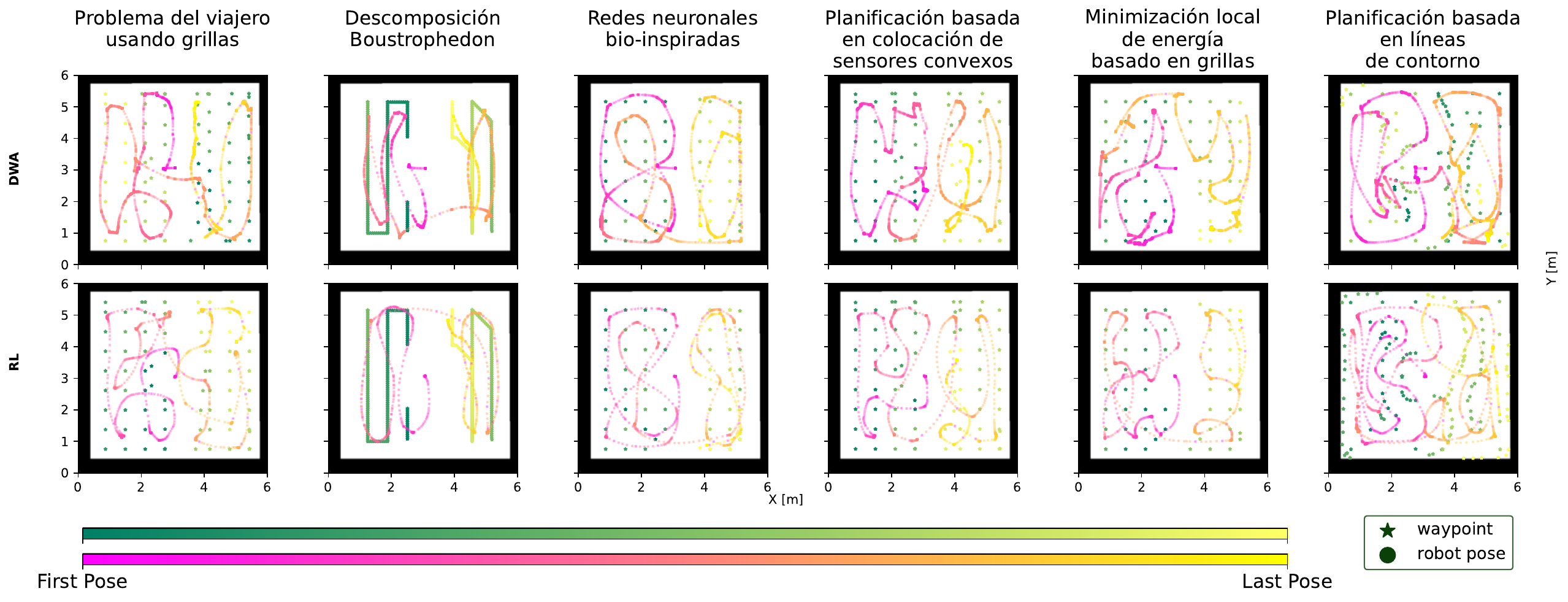}
    \caption{Planes y trayectorias obtenidas en experimento en escenario de prueba simple en simulación.}
    \label{fig:exp-simu-simple}
    \vspace{-10pt}
\end{figure}

En cuanto a la eficacia de los algoritmos de planificación de rutas de cobertura, se observan tendencias similares a las vistas en el \textit{benchmark}, con coberturas comparables para los algoritmos con mejores resultados. Para los algoritmos con menor desempeño, el de descomposición Boustrophedon mostró un resultado dentro del rango esperado según la Tabla~\ref{tab:benchmark}. En cambio, el algoritmo de planificación basado en líneas de contorno obtuvo un resultado significativamente mejor que en el obtenido en el \textit{benchmark}. Como se suponía en la sección anterior, aunque la densidad de las poses no es muy alta en este caso, su disposición permite que, si el robot sigue una ruta recta entre poses consecutivas, se logre una cobertura elevada.  La Figura \ref{fig:exp-simu-simple} permite comparar visualmente los planes de cobertura y las trayectorias seguidas por el robot, obtenidos usando ambos sistemas de navegación.

En cuanto al desempeño general, se observó dificultad para cubrir los bordes del mapa, y especialmente las esquinas, donde el robot no lograba llegar o demoraba antes de lograr avanzar a la siguiente pose.

\begin{wrapfigure}{r}{0.55\linewidth}
\vspace{-10pt}
    \centering
    \includegraphics[width=1\linewidth]{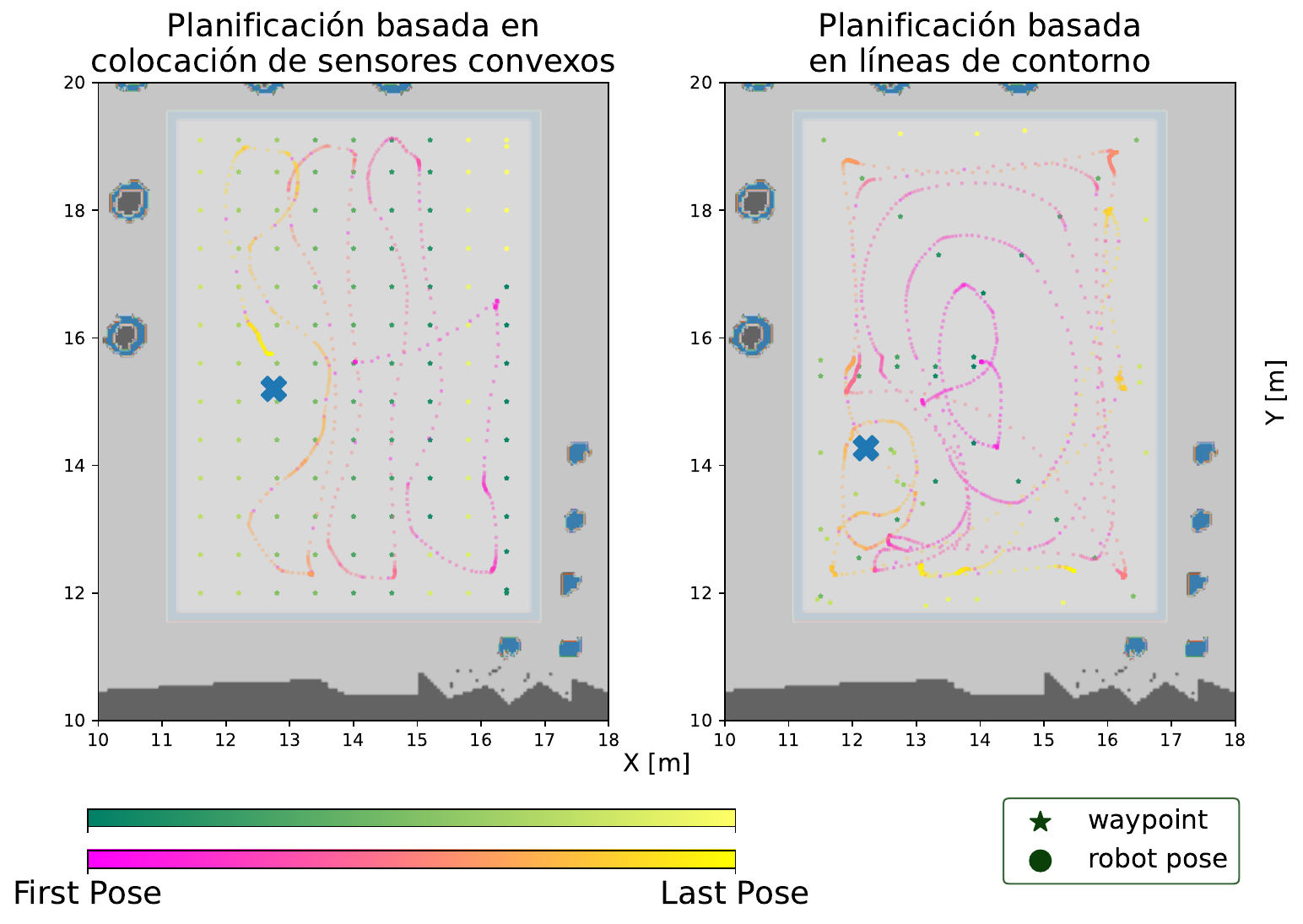}
    \caption{Resultados del experimento con obstáculo no identificado persistente (cruz azul) y planificador local basado en RL.}
    \label{fig:obstaculo-persistente}
    \vspace{-10pt}
\end{wrapfigure}

Para evaluar la capacidad del sistema de re-planificar en áreas que quedaron sin cubrir, se realizó un experimento en el entorno ilustrado por las Figuras~\ref{fig:map_obst_1} y~\ref{fig:map_obst_2}, donde el robot se encontró con un obstáculo no mapeado y persistente. Los resultados en este caso mostraron que, a mayor densidad de poses, la capacidad del sistema para evitar obstáculos disminuye. Esto se debe principalmente al margen reducido para maniobrar y a que algunas poses se encuentran dentro o detrás del obstáculo, lo que hace que el robot intente repetidamente alcanzar un área inaccesible. Esta situación provoca que a veces el robot colisione, como se observa para la planificación basada en colocación de sensores convexos (Figura \ref{fig:obstaculo-persistente}, izquierda). En el caso de \texttt{move\_base}, se observó que al enfrentarse a un obstáculo el robot tiende a quedarse estancado, no pudiendo la mayoría de las veces seguir con la tarea. No obstante lo anterior, en los casos en que el robot logra completar la tarea evitando el obstáculo, se observó que el sistema puede re-planificar para cubrir las áreas que quedan sin cubrir inicialmente, como se muestra en el caso del algoritmo de planificación basado en líneas de contorno (Figura \ref{fig:obstaculo-persistente}, derecha). Es importante destacar que estas áreas sin cubrir deben ser al menos del tamaño del radio de cobertura del robot para que los planificadores puedan re calcular. Esta es una limitación del sistema que deberá abordarse en el futuro. 

\vspace{-10pt}
\subsubsection*{Pruebas en mundo real}
\begin{figure}[H]
    \vspace{-10pt}
    \centering
    \includegraphics[width=1\linewidth]{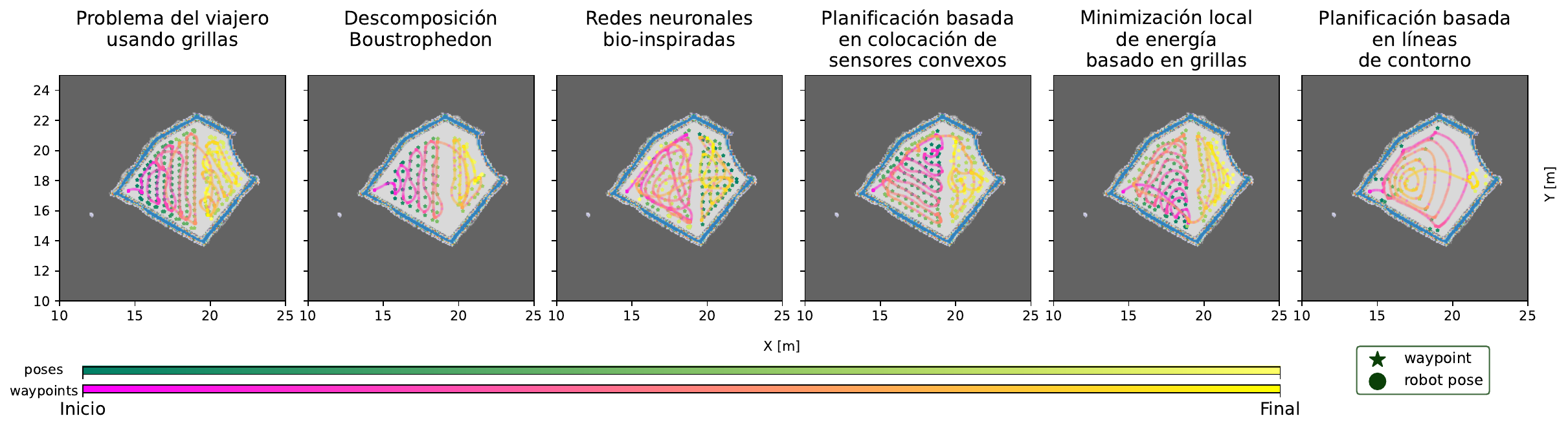}
    \caption{Planes y trayectorias obtenidas en experimento en escenario de prueba sub-sección simple pileta AMTC.}
    \label{fig:pileta-amtc-slice}
    \vspace{-5pt}
\end{figure}

\begin{wraptable}{r}{0.55\linewidth}
\vspace{-14pt}
\caption{Resultados prueba sub-sección simple pileta AMTC.}
\label{tab:pileta-amtc-slice}
\resizebox{0.55\columnwidth}{!}{
\begin{tabular}{lcc}%ccc
\toprule
\textbf{Algoritmo}&  \textbf{$C_{R\%}$ [\%]} & $T_{\text{exec}}$ \textbf{[min]} \\ \midrule
Problema del viajero usando grillas & 96.80 & 4.90  \\
Descomposición Boustrophedon & 87.34 & 3.16 \\
Redes neuronales bio-inspiradas & 96.05 & 5.35 \\
P. basada en colocación de sensores convexos & 96.72 & 4.65 \\
Minimización local de energía basada en grilla & 95.53 & 4.95 \\
\bottomrule
\end{tabular}}
\end{wraptable}

El experimento realizado usando la delimitación de la Fig.~\ref{fig:slice_pileta} permite observar el comportamiento del sistema en el mundo real. Este primer entorno simple utilizado facilita la comparación con lo visto en simulación, no obstante, dados los altos tiempos de ejecución de la tarea medidos para DWA (ver Tabla~\ref{tab:cuadrado-simu}), esta vez solo el planificador local basado en RL es utilizado para evaluar el sistema.

Los resultados asociados, presentados en la Figura~\ref{fig:pileta-amtc-slice} y en la Tabla~\ref{tab:pileta-amtc-slice}, son similares a los obtenidos en el experimento en simulación de la Tabla~\ref{tab:cuadrado-simu}, mostrando una leve mejora en la cobertura para todos los algoritmos, excepto en la descomposición Boustrophedon, que experimentó una disminución. Además, se observó una mejora en el tiempo de ejecución de la tarea, que fue menor que en simulación, a pesar de que el entorno era ligeramente más extenso.

\begin{figure}[H]
    \centering
    \includegraphics[width=1\linewidth]{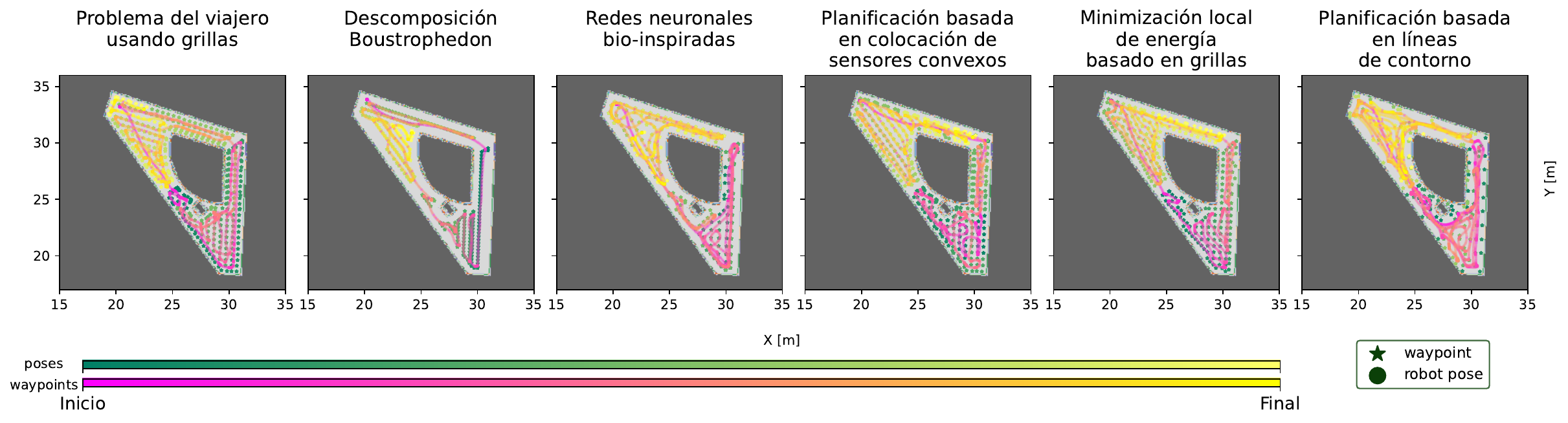}
    \caption{Planes y trayectorias obtenidas en experimento en escenario de prueba circuito pileta AMTC.}
    \label{fig:pileta-amtc-circuit}
    \vspace{-10pt}
\end{figure}

\begin{wraptable}{r}{0.6\linewidth}
\vspace{-12pt}
\caption{Resultados prueba en circuito pileta AMTC. }
\label{tab:pileta-amtc-circuit}
\resizebox{0.55\columnwidth}{!}{
\begin{tabular}{lcc}
\toprule
\textbf{Algoritmo}&  \textbf{$C_{R\%}$ [\%]} & $T_{\text{exec}}$ \textbf{[min]} \\ \midrule
Problema del viajero usando grillas & 89.87 & 11.48  \\
Descomposición Boustrophedon & 68.41 & 5.40 \\
Redes neuronales bio-inspiradas & 79.20 & 11.15 \\
P. basada en colocación de sensores convexos & 90.13 & 12.06 \\
Minimización local de energía basada en grilla & 83.74 & 8.86 \\
Planificación basada en lineas de contorno & 85.45  &  17.26 \\
\bottomrule
\end{tabular}}
%\vspace{-10pt}
\end{wraptable}

\label{exp:pileta-amtc}
Un último experimento fue realizado en un ambiente más desafiante, ilustrado en la Fig.~\ref{fig:map_pileta}, donde se incorpora un obstáculo en medio del área a cubrir (la pileta mostrada en la Fig~\ref{fig:map_real}). Los resultados son resumidos en la Tabla \ref{tab:pileta-amtc-circuit} y en la Figura \ref{fig:pileta-amtc-circuit}, donde se pueden observar los planes generados en contraste a la trayectoria ejecutada por el robot. A partir de los resultados obtenidos, se observa que en este caso hubo una baja importante en el desempeño del sistema, pudiendo alcanzarse una cobertura cercana al 90\% en dos casos, los cuales corresponden a los algoritmos que alcanzaron mayor cobertura promedio en el \textit{benchmark}. Esta baja en la cobertura de todos los algoritmos, nuevamente puede ser explicada por la dificultad que tiene el sistema para cubrir bordes y esquinas, las que en este caso son numerosas al haber un obstáculo de gran tamaño en el medio del área a cubrir. Para los algoritmos con bajas más significativas, como descomposición Boustrophedon y redes neuronales bio-inspiradas, este detrimento en desempeño podría indicar que en casos donde hay presencia de obstáculos, estos algoritmos podrían no ser adecuados para un sistema como el presentado en este trabajo, principalmente por limitaciones asociadas al planificador local usado.

\section*{DISCUSIÓN Y CONCLUSIONES}

El sistema presentado en este trabajo dota a una plataforma móvil con la capacidad de cubrir un área, permitiendo que la base móvil planifique y navegue de forma autónoma, evadiendo obstáculos y re-planificando para cubrir secciones que hayan quedado sin cobertura.

El uso de diferentes algoritmos de planificación de rutas de cobertura genera distintos patrones de movimiento en el robot, desde planes simples, como los de la descomposición Boustrophedon, hasta más complejos, como los generados por el algoritmo de redes neuronales bio-inspiradas. Además, se observó cómo el planificador local influye tanto en el porcentaje de cobertura alcanzada como en el tiempo de ejecución de la tarea, siendo por lo tanto una componente esencial para el sistema.

Aunque en general se obtuvieron buenos resultados, con un alto porcentaje de área cubierta en casi todas las pruebas, se identificaron debilidades y oportunidades de mejora. Principalmente, el sistema muestra dificultades para cubrir los bordes del mapa, tanto en los límites exteriores como en los de los obstáculos, lo cual afecta negativamente la cobertura final. Esta limitación podría abordarse añadiendo un comportamiento específico que identifique y cubra estas áreas, lo cual se espera que mejore significativamente el desempeño del sistema, dado que la mayor pérdida de porcentaje de cobertura se debió a este problema.

Otra debilidad del sistema es que los planificadores de rutas de cobertura no logran encontrar un plan cuando las áreas a cubrir son más pequeñas que el radio de cobertura de la plataforma móvil. Este problema podría resolverse mediante procesamiento de imágenes, agrandando artificialmente el área sin cubrir hasta que sea lo suficientemente grande para que los planificadores puedan calcular un plan.

Otra posible mejora del sistema está en el planificador local. Para la cobertura, a veces es preferible un planificador que realice movimientos simples, como avanzar en línea recta y girar sobre su eje, sin traslación, cuando la plataforma de despliegue lo permite. Además, es importante que sea capaz de retroceder para salir de atascos en lugares estrechos. También se recomienda que el planificador no intente avanzar si las poses consecutivas están bloqueadas por un obstáculo no identificado previamente, ya que, si el objeto es persistente, el movimiento podría provocar una colisión; en su lugar, la plataforma debería esperar o descartar dichas poses.

Otra oportunidad de mejora en el desempeño del sistema es suavizar los planes de cobertura. En algunos planes calculados se observan \textit{ripples} y poses desordenadas que dificultan su ejecución debido a los cambios bruscos de movimiento que su seguimiento implica, especialmente en espacios con obstáculos y bordes irregulares. Este problema podría abordarse tratando el plan como una señal y aplicando técnicas de suavizado, como el uso de series de Fourier visto en \cite{Tripicchio2023SmoothCP}, para eliminar picos y \textit{ripples}, que corresponden a componentes de alta frecuencia.

\section*{AGRADECIMIENTOS} 
Este trabajo fue parcialmente financiado por el proyecto ANID-PIA AFB230001. Los autores agradecen también a Pablo Alfessi, quien diseñó y construyó la montura para el sensor Ouster, la cual se utilizó para la configuración experimental de simulación y del mundo real. Los autores también agradecen a Gonzalo Olguín, quien implementó una versión preliminar del \textit{raycasting} del sistema.

\bibliographystyle{apacite}
\bibliography{references.bib}

\end{document}